# Harnessing LSTM for Nonlinear Ship Deck Motion Prediction in UAV Autonomous Landing amidst High Sea States

Feifan Yu[1,2,3][0000-0002-3205-0040], Wenyuan Cong[1,2,4][0009-0004-7768-2692], Xinmin Chen[1,2][0000-0003-3278-2185], Yue Lin[1,2][0009-0002-1940-7700] and Jiqiang Wang[1,2*][0000-0002-1317-0880]

[1] Zhejiang Provincial Engineering Research Centre for Special Aircrafts, Ningbo 315336, China
[2] Ningbo Institute of Materials Technology & Engineering, CAS, Ningbo 315201, China
[3] University of Chinese Academy of Sciences, Beijing 101408, China
[4] Faculty of Electrical Engineering and Computer Science, Ningbo University, Ningbo 315211, China
wangjiqiang@nimte.ac.cn

**Abstract.** Autonomous landing of UAVs in high sea states requires the UAV to land exclusively during the ship deck's "rest period," coinciding with minimal movement. Given this scenario, determining the ship's "rest period" based on its movement patterns becomes a fundamental prerequisite for addressing this challenge. This study employs the Long Short-Term Memory (LSTM) neural network to predict the ship's motion across three dimensions: longitudinal, transverse, and vertical waves. In the absence of actual ship data under high sea states, this paper employs a composite sine wave model to simulate ship deck motion. Through this approach, a highly accurate model is established, exhibiting promising outcomes within various stochastic sine wave combination models.

## 1 Introduction

The establishment of a robust maritime nation hinges on a formidable air-sea three-dimensional transportation network. Advancing the capabilities of near-sea Unmanned Aerial Vehicles (UAVs) for tasks such as maritime patrol, search and rescue[1], and emergency transport significantly enhances maritime transportation potential. This mode of transport holds a pivotal role in future development strategies. Nonetheless, the landing phase in naval aviation constitutes merely 4% of the entire flight process, yet contributes to 44.4% of all aviation accidents. Remarkably, unmanned aircraft landing incidents contribute up to 80% of these mishaps. As a result, it becomes imperative to pursue autonomous landing navigation control technology[2] to enhance landing success rates.

Various techniques exist for drone landings, similar to the deck runway landings of carrier-based aircraft[3]. However, one drawback of this approach is the requirement for an extended runway, limiting its suitability primarily to large vessels. The crash net landing[4] method is also prevalent, but it escalates the risk of drone damage and leads to considerable maintenance expenses. Furthermore, sky hook[5] and parachute landings[6] are alternative strategies, albeit susceptible to environmental influences and necessitating precise parachute deployment.

Given the shortcomings of these landing methods, this study opts for the more advantageous vertical take-off and landing approach of tilt-rotor UAVs[7]. This method not only ensures a smooth, non-detrimental interaction with both UAVs and ships, but also demonstrates distinct merits in facilitating UAV landings amid high sea conditions. Specifically, this advantage is underscored by the ship's random six-degree-of-freedom movement and the time allocated for the UAV to adjust its landing point during vertical take-off and landing. This adjustment ensures the ship's tilt remains within an acceptable range upon landing, thereby significantly boosting the success rate of UAV ship landings in challenging sea conditions.

Nonetheless, the vertical take-off and landing method is not exempt from certain limitations. Specifically, these limitations manifest during drone landings, where the ship's inclination due to wave effects must remain within a certain range. Landings are viable only when the ship's tilt remains within an acceptable angle. The time interval during which the ship's tilt falls within this acceptable angle is referred to as the "rest period." Safely landing the drone during this rest period is considered secure. As a result, the realm of autonomous UAV landings necessitates an exploration into software-based predictions of a ship's three-dimensional inclination magnitude induced by wave forces.

Building upon the aforementioned context, this paper employs the Long Short-Term Memory (LSTM)[8] network to prognosticate the three-dimensional motion pattern of a vessel subjected to wave influences. Tailoring its focus to the practical application milieu, this study centers on ship motion within sea state 5 conditions. Given the current unavailability of ship motion data for sea state 5, this research utilizes Huang's[9] sine wave superposition methodology to replicate the deck motion of Knox-class warships, subsequently leveraging this synthesized dataset for model training. Subsequently, a stochastic sine wave combination model is constructed based on sea state data at level 5 to validate the precision of the proposed model presented in this study.

The forthcoming thesis can be broadly divided into four parts. Chapter 2 commences by outlining the dataset construction procedure, followed by an exposition on the significance of each individual indicator. In Chapter 3, the LSTM model is employed. Initially, the model combines indicators across three dimensions and subsequently applies this amalgamation to predict the previously mentioned dataset, ultimately yielding a high-precision predictive model. In Chapter 4, a stochastic sine wave composite model is constructed to assess the trained model's performance. This evaluation is accomplished by utilizing known sea state parameters corresponding to the 5-level classification. The aim is to affirm the model's precision and its comparative advantages. Chapter 5 encapsulates the content discussed thus far, concluding with a summary of the findings and offering insights into prospective avenues for future research.

## 2   Data preparation

When a ship traverses the sea under conditions of high sea state, it becomes susceptible to substantial wind and wave influences. Consequently, the ship experiences intricate six-degree-of-freedom movements[10]. Within the context of UAV ship landings, the attainment of relative hovering between the UAV and the ship has been achieved. As a result, it becomes feasible to exclude the longitudinal and transverse lateral displacement components during the deconstruction of this intricate six-degree-of-freedom motion. Consequently, we decompose this intricate six-degree-of-freedom motion into three distinct modes of movement: the ship's longitudinal, transverse, and vertical oscillations.

The longitudinal and transverse oscillations pertain to the ship's amplitude of movement along its longitudinal or transverse axes due to wind and wave forces. Conversely, vertical oscillations denote alterations in the ship's vertical displacement as a result of wind and wave influences. This is visually illustrated in the figure presented below.

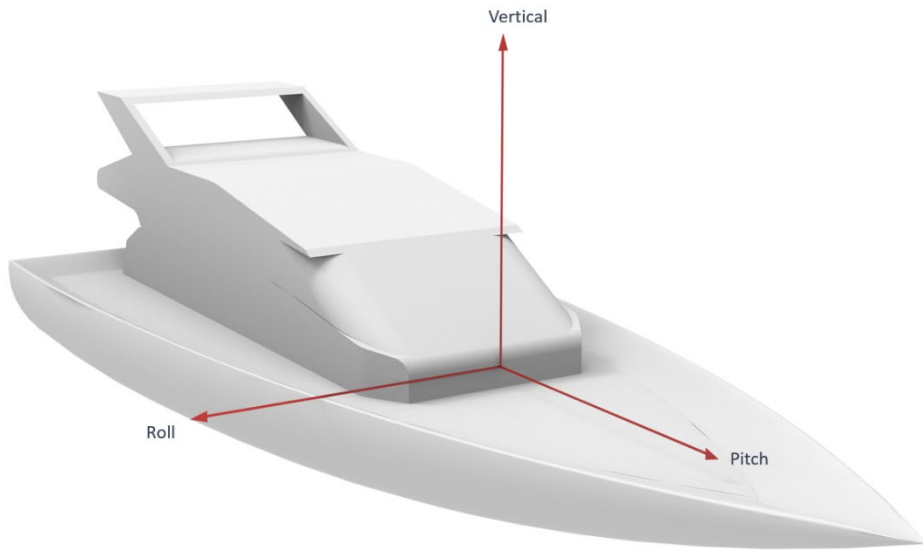

Figure 1. Illustration of the ship rocking model

Given the absence of three-dimensional ship motion data under the impact of waves during high sea states, this study employs the sine wave superposition methodology outlined in Huang's work to replicate the deck movement of Knox-class warships. The simulated model for combining sine waves is detailed as follows:
$$h_s(t) = 0.2172\sin(0.4t) + 0.4714\sin(0.5t) + 0.3592\sin(0.6t) + 0.2227\sin(0.7t)$$

$$\theta_s(t) = 0.005\sin(0.46t) + 0.00946\sin(0.58t) + 0.00725\sin(0.7t) + 0.00845\sin(0.82t)$$
$$\varphi_s(t) = 0.021\sin(0.46t) + 0.0431\sin(0.54t) + 0.029\sin(0.62t) + 0.022\sin(0.67t)$$

In this model, $\theta_s(t)$ represents the longitudinal swing, $\varphi_s(t)$ signifies the transverse swing, and $h_s(t)$ denotes the vertical swing. The measurements for $\theta_s(t)$ and $\varphi_s(t)$ are expressed in terms of swing inclination units, while $h_s(t)$ represents the height of elevation or descent. The visual depiction of this configuration is illustrated below:

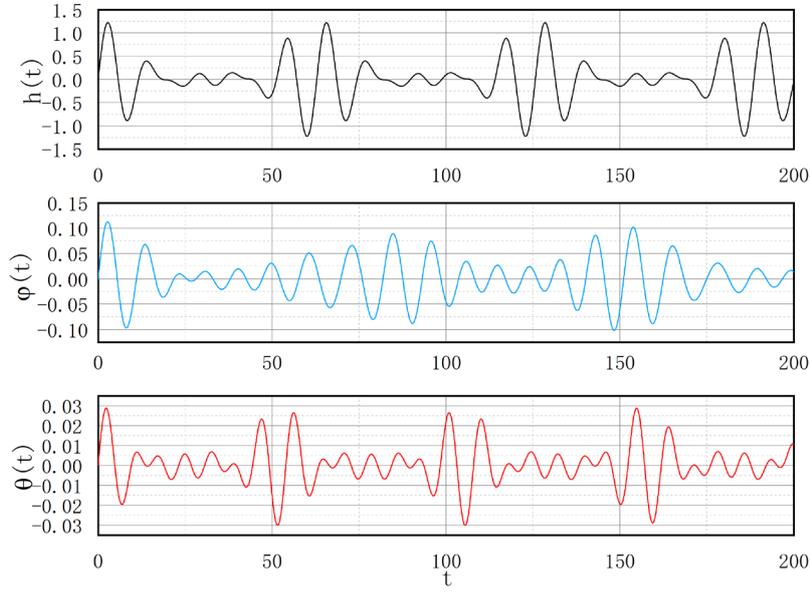

Figure 2. sinusoidal combinatorial model

## 3 Model training

In this section, we will proceed to train the LSTM model with the aim of predicting three key parameters of a ship's behavior under wave influence: the longitudinal swing, the angle of the transverse swing, and the height of the vertical swing.

Initially, we divide each of the sinusoidal wave combination models mentioned above into sets of 2000 data points. These data point sets are subsequently employed as both our training and testing datasets. The initial 70% of data points are designated as the training set, while the remaining 30% are allocated for testing purposes.

The crux of our training approach hinges on harnessing the inherent capabilities of LSTM. This involves predicting the magnitude of the forthcoming data point based on the learning of 40 consecutive data points leading up to the predicted data point. This process is reiterated to predict the entire curve depicting the ship's motion attitude.

Furthermore, the model incorporates a composite LSTM neural network prediction architecture that concurrently forecasts three distinct output parameters—namely, the longitudinal, transverse, and vertical sways—as a unified prediction. This feature notably underscores the paper's strengths.

For model training, a continuous dataset comprising 2000 data points is employed. Precisely, the initial 70% of the dataset, equating to 1400 points, constitutes the training set used to train our network. Subsequently, the remaining 30% (600 points) is allocated for testing the predictions generated by our network. The outcomes of the training and testing sets are depicted in Figures 3 and 4.

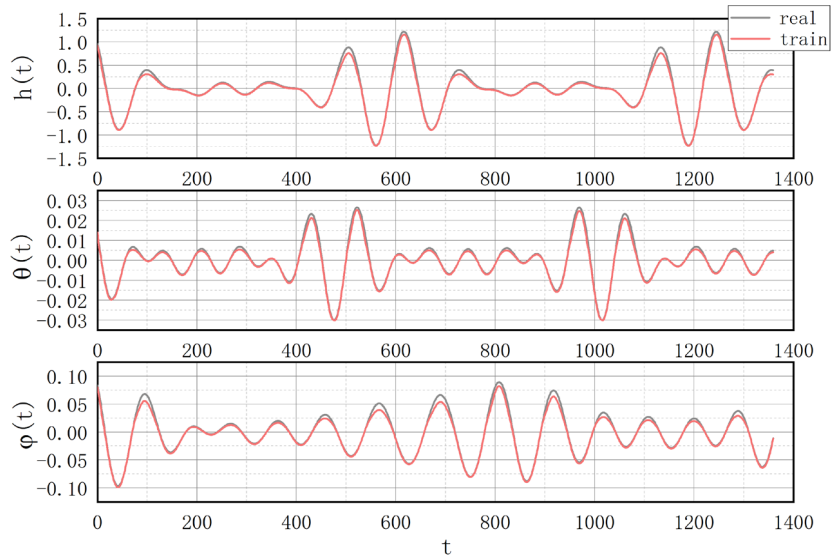

Figure 3. training process training set effect

In Figure 3, it is evident that the current model exhibits a near-flawless performance in predicting the transverse swing $\varphi_s(t)$ curve on the training set, demonstrating minimal errors. Although the outcomes for the longitudinal swing $\theta_s(t)$ curve and vertical swing $h_s(t)$ curve are marginally less remarkable compared to the $\varphi_s(t)$ curve, they nevertheless remain highly satisfactory.

Figure 4 illustrates that the current model adeptly forecasts all three curves on the test set. However, it's noteworthy that while the prediction for the $h_s(t)$ curve remains highly accurate, there is a relatively minor jitter observed in its performance during the initial 300 data points of the test set. It's important to highlight that through experimentation, it was observed that increasing the number of hidden neurons in the LSTM correlates with a reduction in the amplitude of this jitter, consequently enhancing the prediction quality.

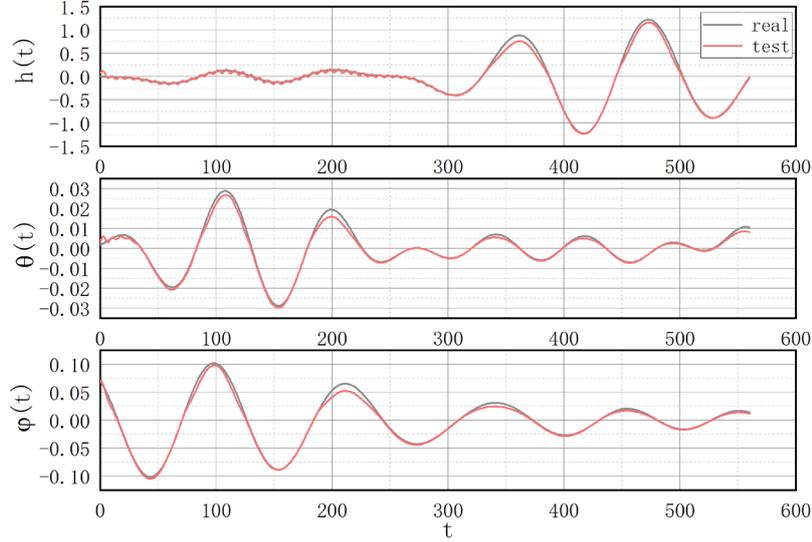

Figure.4. Training process test set effect

## 4    Model validation

The foregoing highlights the exceptional performance of this model across both the training and test sets. The ensuing section proceeds to examine the model's efficacy in the context of high sea state conditions. Specifically, this study focuses on the validation and generalization capabilities of the model within a class 5 sea state scenario. This particular sea state is characterized by a significant wave height ranging from 2.5 to 4 meters and a characteristic period spanning 5.5 to 6.7 Hz.

Initial steps involve the random generation of a sinusoidal combination model to emulate the deck movement of a ship within a class 5 sea state. A few of the noteworthy parameters characterizing the class 5 sea state are presented in the subsequent table:

Table 1. Selected reference data for sea state 5

|  | amplification | cyclicality |
|---|---|---|
| $h_s(t)$ | $1.0 < \bar{Z}_{1/3} < 1.9$ | $5 < T < 8$ |
| $\varphi_s(t)$ | $6.3 < (\bar{\varphi}_a)_{1/3} < 12.0$ | $8 < T < 13$ |
| $\theta_s(t)$ | $1.3 < (\bar{\theta}_a)_{1/3} < 2.5$ | $5 < T < 8$ |

Utilizing the data provided in Table 1, we embark on the random generation of multiple sets of sine wave combination models. Upon observation, it becomes apparent that the outcomes are consistently congruent. Hence, this paper selects a representative set from these outcomes to effectively illustrate the predictive prowess of the current model within the context of class 5 sea state. The composite sine wave model is depicted as follows:

$h_s(t) = 0.25\sin(0.785t) + 0.35\sin(0.9t) + 0.45\sin(1.1t) + 0.5\sin(1.256t)$

$\theta_s(t) = 0.35\sin(0.8t) + 0.45\sin(0.85t) + 0.55\sin(0.95t) + 0.625\sin(1.156t)$

$\varphi_s(t) = 2.6\sin(0.483t) + 1.8\sin(0.5t) + 2.5\sin(0.6t) + 3\sin(0.785t)$

The precise waveform impact is graphically demonstrated in Figure 5. A noteworthy observation is the substantial amplification in amplitude and frequency of the $h_s(t)$, $\theta_s(t)$, and $\varphi_s(t)$ waveforms under the influence of the five levels of sea state, when compared to the waveforms utilized in the earlier training model.

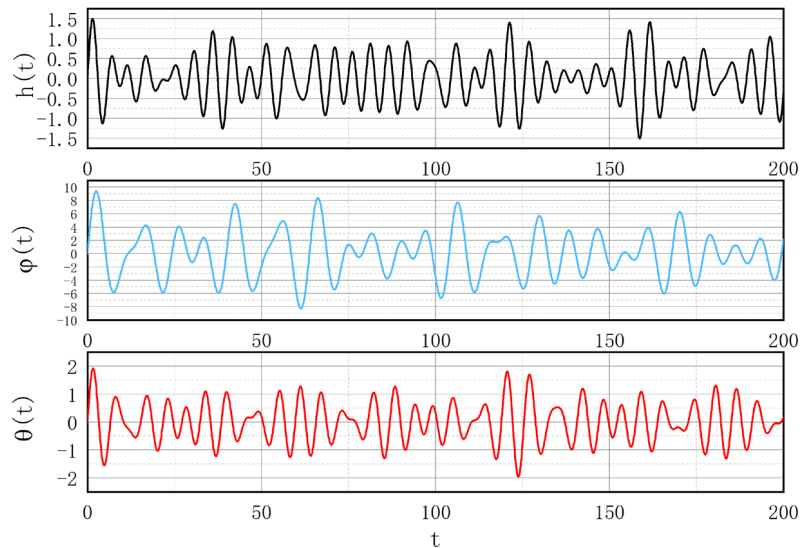

Figure.5 Combined sine wave model for five levels of sea state

Subsequently, the LSTM neural network composite model, trained as described earlier, was employed to forecast the curves depicting deck motion under the influence of class V sea state. These curves were generated through the stochastic sine wave combination model.

The dataset still comprises 2000 data points, as each prediction of a data point is deduced from the preceding 40 data points. In other words, the forecast for the 41st data point relies on information from the 1st through the 40th data points. Consequently, the 41st data point marks the inaugural prediction of this model, and this sequential process extends to the 2000th data point. Therefore, a total of 1960 data points are predicted by this model.

Given the inherent correlation among the curves, this paper has opted not to employ three separate LSTM neural networks for forecasting the longitudinal, transverse, and vertical swing curves. Instead, a unified LSTM neural network is utilized to predict the composite of these three curves. The objective is to concurrently unravel the latent interdependencies among the individual curves, enhancing the model's predictive capabilities.

This study employs an LSTM composite neural network to forecast the motion attributes of a ship's deck at the fifth sea state level, as illustrated in Figure 6. As evident from the figure, the prediction of the transverse swing curve is executed with remarkable precision, yielding an exceedingly minute error margin. Although the predictive accuracy for the longitudinal and vertical sway curves might not attain the same level as that of the transverse sway, the projected curves continue to uphold a fundamental alignment with the original curves.

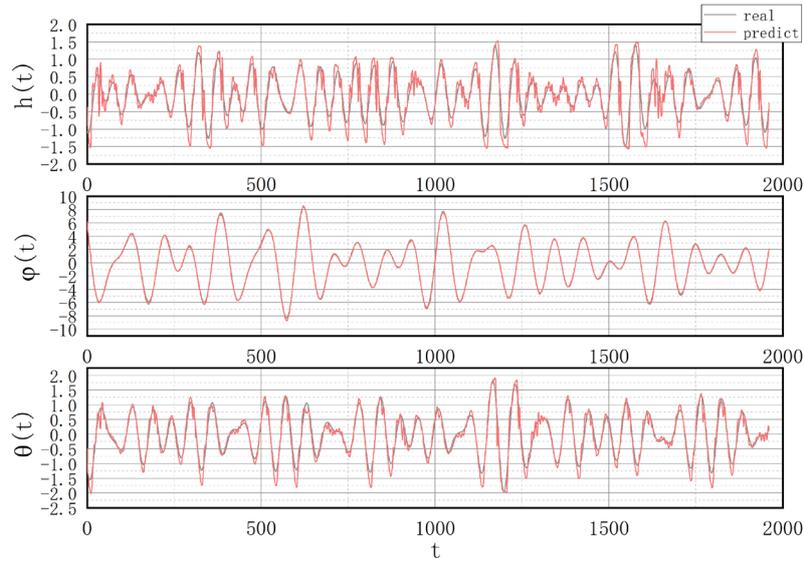

Figure.6 Effectiveness of LSTM prediction for five levels of sea state

As the contours of the original curve and the predicted curve exhibit fundamental congruence, this study has introduced absolute error curves[11] for both the original and predicted curves, as depicted in Figure 7. This has been done to provide enhanced clarity in assessing the extent of the errors.

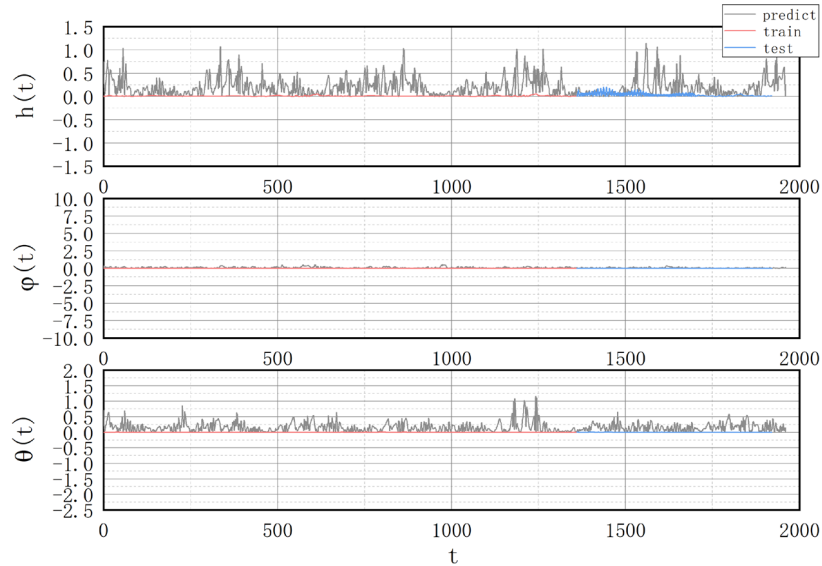

Figure 7. Absolute error between original and predicted plot lines

From the figure, it's apparent that the prediction error for the ship's roll motion is exceptionally minimal. This observation underscores the robust generalization capability of the model trained in this study for forecasting roll motion. Similarly, although the prediction error for ship's pitch motion might not achieve the same level of perfection as roll motion, it remains notably slight.

However, with respect to the ship's heave motion, while the predicted contour generally aligns with the original curve, a significant proportion of its amplitudes marginally exceed those of the original. In practical terms, since the predicted heave motion amplitude is only slightly elevated, it doesn't significantly impede the prediction of the "resting period" essential for drone landing. Consequently, the overall predictive performance for heave motion remains considerably robust.

It is reasonable to assume that the present capability for predicting deck motion is largely satisfactory and can serve as a pivotal technology for the preliminary stages of autonomous UAV landing. Naturally, the focal point of subsequent endeavors lies in refining this prediction by reducing errors and enhancing the accuracy of the LSTM neural network harnessed in this context. This refinement is poised to play a pivotal role in the advancement of future work.

## 5 Summary

This paper introduces several noteworthy innovations:

1) The paper introduces a pioneering methodology to achieve autonomous landing for unmanned aerial vehicles. Specifically, it harnesses LSTM neural networks to predict the three-dimensional motion patterns of ship decks. This serves as a foundational information source, enabling subsequent efforts to identify "resting periods" and facilitate the smooth landing of unmanned aerial vehicles during these periods.

2) The paper innovatively amalgamates the roll, pitch, and heave motions of ships into a singular LSTM neural network model for prediction. Firstly, this composite model replaces the need for three separate models, leading to a substantial reduction in model complexity. Secondly, the model accounts for the interconnections between the three modes of movement, further enhancing its predictive capabilities.

The model proposed in this paper is not without limitations. While it demonstrates commendable performance in predicting pitch and roll curves across both the training and test sets, its predictive accuracy of actual values is marginally compromised. This could potentially be attributed to the application of a random sine wave combination model within this study to simulate real-world sea state level 5 conditions. Within a random environmental context, the inherent interplay among the three modes of motion in the sine wave combination model might be diluted or disregarded. Furthermore, the scarcity of training data poses a challenge, as the majority of the available data corresponds to sea state levels 2 to 3. The absence of level 5 sea state data for model training introduces certain limitations and contributes to the observed errors.

As we look ahead to future endeavors, it becomes apparent that further work is essential. Primarily, there is a critical need to venture into the open seas and amass authentic 3D motion data of ship decks across sea state levels 1 to 5. This real-world dataset would serve to validate the efficacy of our model and, in turn, facilitate the development of more precise neural network models derived from this expanded dataset.

Turning our attention to the model itself, the distinct intricacies of autonomous landing projects for unmanned aerial vehicles in high sea conditions necessitate a continuous process of refinement and debugging. This iterative approach is pivotal in achieving the highest levels of accuracy, thereby supplying the requisite informational foundation for ensuring the seamless landing of unmanned aerial vehicles.